\documentclass[review]{elsarticle}

\usepackage{lineno,hyperref}

\journal{Journal of \LaTeX\ Templates}

\usepackage{algorithm}
\usepackage{algorithmic}
\usepackage{subfigure}
\usepackage{amssymb}
\usepackage{multirow}
\usepackage{bbding}
\usepackage{color}

\journal{Information Fusion}

\begin{document}
\begin{frontmatter}

\title{Multi-view Information Integration and Propagation for Occluded Person Re-identification}

\author[]{Neng Dong$^{a}$}
\author[]{Shuanglin Yan$^{a}$}
\author[]{Hao Tang$^{a}$}
\author[]{Jinhui Tang$^{a}$}
\author[]{Liyan Zhang$^{b}$\corref{mycorrespondingauthor}}
\address{a. School of Computer Science and Engineering, Nanjing University of Science and Technology, Nanjing 210094, P.R. China}
\address{b. College of Computer Science and Technology, Nanjing University of Aeronautics and Astronautics, Nanjing 210016, P.R. China}

\begin{abstract}
Occluded person re-identification (re-ID) presents a challenging task due to occlusion perturbations. Although great efforts have been made to prevent the model from being disturbed by occlusion noise, most current solutions only capture information from a single image, disregarding the rich complementary information available in multiple images depicting the same pedestrian. In this paper, we propose a novel framework called Multi-view Information Integration and Propagation (MVI$^{2}$P). Specifically, realizing the potential of multi-view images in effectively characterizing the occluded target pedestrian, we integrate feature maps of which to create a comprehensive representation. During this process, to avoid introducing occlusion noise, we develop a CAMs-aware Localization module that selectively integrates information contributing to the identification. Additionally, considering the divergence in the discriminative nature of different images, we design a probability-aware Quantification module to emphatically integrate highly reliable information. Moreover, as multiple images with the same identity are not accessible in the testing stage, we devise an Information Propagation (IP) mechanism to distill knowledge from the comprehensive representation to that of a single occluded image. Extensive experiments and analyses have unequivocally demonstrated the effectiveness and superiority of the proposed MVI$^{2}$P. The code will be released at \url{https://github.com/nengdong96/MVIIP}.
\end{abstract}

\begin{keyword}
Occluded person re-ID, multi-view information, information integration, information propagation.
\end{keyword}

\end{frontmatter}


\section{Introduction}

Person re-identification (re-ID) aims to retrieve specific pedestrians from a large-scale cross-camera image database. This field has garnered significant interest due to its practical implications in various intelligent security applications \cite{SHABAN, HIGCN, CCG-LSTM, Yan1, Yan2}. However, most mainstream methods \cite{AN, SUUTALA, AGW, BAGTRICKS} rely on an ideal assumption that the complete body of each pedestrian is visible for identification, disregarding situations where the target pedestrian may be obstructed by obstacles or other individuals in real-world scenarios. To address this limitation, occluded person re-ID \cite{OCCLUDEDREID} has been proposed, which retrieves images with the same identity as the given occluded query from a gallery where both holistic and occluded images are present.

\begin{figure}[th!]
\centering
\subfigbottomskip=-1pt
\subfigcapskip=-1pt
{\includegraphics[height=2.4in,width=3.4in,angle=0]{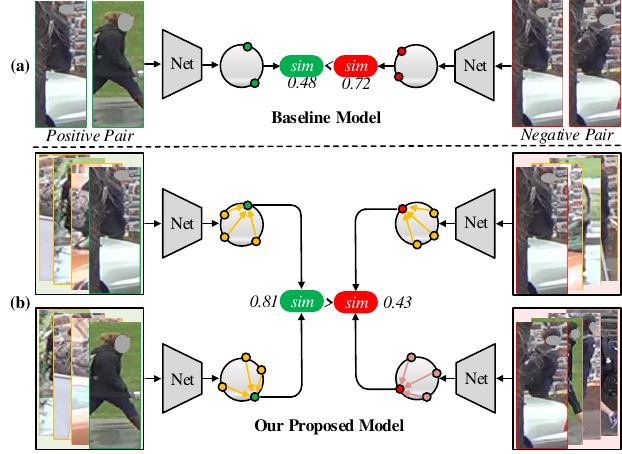}}
\caption{The core motivation of this paper. (a) Existing methods (i.e., the baseline model) extract features independently from single-view images to match identity, resulting in the similarity of pedestrian images with different identities greater than that of pedestrian images with the same identity. (b) Multiple images with the same identity provide rich complementary information. Integrating such multi-view information holds the potential to comprehensively characterize the occluded target pedestrian, correctly distinguishing pedestrians.}
\label{Fig:1}
\end{figure}

Although great efforts \cite{PGFA, HOREID, OPRDAAO} have been made to prevent the model from being disturbed by occlusion noise, current solutions predominantly capture information from a single image, disregarding the rich complementary one available in multiple images depicting the same pedestrian, which still hinders these methods from accurately matching identity. To be more specific, most of the trained models fail to effectively characterize the occluded target pedestrian as the information presented in visible areas of a single image is inherently limited. In particular, as illustrated in Figure \ref{Fig:1} (a), the visible regions of the same pedestrian may vary, while the visible regions of different pedestrians may share a similar appearance. Existing frameworks extract features independently from each individual image to measure similarity, posing challenges in distinguishing identities. In general, as shown in Figure \ref{Fig:1} (b), there are multiple images depicting the same pedestrian, and visible body parts across different images may exhibit substantial variation, providing comprehensive information crucial for alleviating the difficulty of identity matching in occluded scenes. Consequently, the core motivation of this paper is to fully exploit the information available in multi-view images, learning a comprehensive representation that accurately captures the characteristics of the occluded target pedestrian.

A straightforward approach to accomplish the aforementioned objective is to integrate the representations of multiple images. However, two crucial considerations arise: 1). Not all information, such as obstacles and non-target pedestrians, is beneficial for identification. Indiscriminately integrating all information from each image may introduce undesirable noise. 2). The contributions to accurate identification vary across different types of information. Equally integrating the information from different images may potentially compromise the overall reliability of the comprehensive representation. Therefore, the primary challenges addressed in this paper revolve around guiding the model to selectively integrate information valuable for identification and emphatically integrate highly reliable information. Additionally, multiple images of the same pedestrian are solely accessible during training. In the inference phase, the query is a single-view image, and it is unknown whether different images belong to the same identity. Therefore, another challenge in this paper lies in endowing the model with the ability to reason multi-view information from a single image.

This paper presents a framework called Multi-view Information Integration and Propagation (MVI$^{2}$P). Specifically, realizing the significant potential of multi-view images in addressing occluded person re-ID, we integrate their feature maps to generate a comprehensive representation. During this process, to prevent the introduction of noisy information, we develop a CAMs-aware Localization module. This module calculates class activation maps (CAMs) to highlight the spatial discriminative regions of the feature maps, with the assistance of which we can selectively integrate information contributing to the identification. Furthermore, considering the varying discriminative nature of different information, we design a probability-aware Quantification module. This module assigns weights to feature maps based on their probability of being correctly classified, prompting the comprehensive representation to be more reliable. Finally, as multiple images with the same identity are not accessible in the testing stage, we introduce an Information Propagation (IP) mechanism, which propagates information from the comprehensive representation to individual image representations through knowledge distillation. Notably, the Localization and Quantification modules do not necessitate additional networks, and the entire framework introduces only an additional classifier, simple yet effective.

Our main contributions are summarized as follows:
\begin{itemize}
\item The developed MVI$^{2}$P framework leverages the wealth of complementary information available in multiple images depicting the same identity, a crucial aspect overlooked in existing occluded person re-ID approaches. By employing MVI$^{2}$P, our model is capable of excavating comprehensive information from a single occluded image.

\item The designed Localization and Quantification modules encourage the model to spontaneously explore the salient regions of individual feature maps and measure their relative importance, selectively and preferentially integrating valuable and reliable information.

\item The experiments conducted on five public datasets offer compelling evidence showcasing the superiority of MVI$^{2}$P. Furthermore, the model analysis validates the effectiveness of each module and highlights MVI$^{2}$P's advantage in terms of model complexity.
\end{itemize}

The remainder of this paper is organized as follows: We first review the related works in Section \uppercase\expandafter{\romannumeral2}; And then, we present the details of the proposed method in Section \uppercase\expandafter{\romannumeral3}; Next, we introduce the experimental comparison and analysis in Section \uppercase\expandafter{\romannumeral4}; Finally, we provide a brief conclusion in Section \uppercase\expandafter{\romannumeral5}.

\section{Related Works}

\subsection{Holistic Person Re-ID}
According to the way of learning the features, existing holistic person re-ID approaches can be broadly categorized into hand-crafted feature-based methods and deep learning-based methods.

Hand-crafted feature-based methods \cite{ELF, SCN, GOG} focus on manually designing features that match human visual cognition, capturing color, texture, and fine-grained information to represent pedestrians. For instance, Gray et al. \cite{ELF} modeled features on color and texture channels to capture viewpoint invariance in pedestrian representation. Yang et al. \cite{SCN} introduced a novel salient color descriptor, estimating distributions in different color spaces to generate pedestrian representations accordingly. Additionally, Matsukawa et al. \cite{GOG} developed a descriptor based on hierarchical pixel distributions to acquire local information. Recently, combining dictionary learning further enhances the performance of the hand-crafted feature-based method \cite{AIESS}. However, designing such features requires careful attention and becomes challenging when dealing with large-scale databases.

Deep learning-based methods \cite{DML, AANet, CamStyle} have garnered significant attention recently for deep neural networks' success in various computation vision tasks \cite{TANG1, TANG2, Boosting, KGS, PLMSA}. Yi et al. \cite{DML} devised a deep re-ID framework with a siamese structure to capture the association of the same pedestrian across different views. Leveraging attributes, Tay et al. \cite{AANet} obtained multi-level attribute information from the human body and used it to construct an attention map for extracting fine-grained discriminative features. Addressing style variations between cameras, Zhong et al. \cite{CamStyle} jointly trained the re-ID model using images generated by CycleGAN \cite{CycleGAN} and the original inputs, enabling learning of a camera-invariant descriptor subspace. However, these methods overlook the prevalent occlusion problem encountered in real-world scenes, resulting in limited practicality of the models. In contrast, the developed MVI$^{2}$P framework tackles the occlusion problem and offers applicability to holistic person re-ID.

\subsection{Occluded Person Re-ID}
According to the utilization of external visual cues, existing occluded person re-ID approaches can be categorized into external model-assisted methods and external model-free methods.

\textbf{External model-assisted methods} \cite{PGFL-KD, SORN, PIRT, OAMN} utilize external tools to identify the non-occluded regions of pedestrians, enabling the re-ID network to extract identification-beneficial features. Zheng et al. \cite{PGFL-KD} employed pose information to construct a foreground-enhanced module and a body part semantics-aligned module. The former addresses obstacle interference, while the latter tackles position misalignment. Zhang et al. \cite{SORN} proposed a semantic-aware network that leverages human segmentation to learn occlusion-robust representations at both global and local levels. Ma et al. \cite{PIRT} developed a pose-guided inter- and intra-part relational transformer to capture information from different visible regions. To address the domain gap faced by the auxiliary model in person re-ID, Chen et al. \cite{OAMN} introduced an attention-guided mask module for precise localization of human body parts. However, these methods rely heavily on external visual cues, which can be mispredicted in complex scenes. In contrast, our MVI$^{2}$P framework overcomes the occlusion problem without requiring external visual cues.

\textbf{External model-free methods} \cite{ISP, ASAN, MHSANET} have garnered significant attention due to their independence from external tools. These methods aim to drive the model to learn discriminative representations spontaneously in visible regions. Zhu et al. \cite{ISP} employed cluster analysis on pedestrian feature maps to generate pseudo-labels for human semantics, enabling the grouping of image pixels into foreground and background. Jin et al. \cite{ASAN} introduced a visible region-matching algorithm to filter out interference information in images, thereby purifying pedestrian features. Additionally, Tan et al. \cite{MHSANET} incorporated a multi-head self-attention module to regularize the learning of global features, enabling the network to adaptively capture essential local information. Recently, Wang et al. \cite{QPM} predicted part quality scores and combined them with an identity-aware spatial module to obtain high-quality pedestrian representations. Zhao et al. \cite{IGOAS} proposed a novel data augmentation technique tailored for occluded re-ID, introducing an incremental generative occlusion block to generate occlusion samples for training. However, we have observed that the prominent algorithms in this category primarily emphasize the extraction of local features and the generation of occlusion samples, resulting in high computational costs. More importantly, these methods independently learn pedestrian representations from individual images, overlooking the wealth of complementary information that multiple images can provide for characterizing the target pedestrian. In contrast, our MVI$^{2}$P framework harnesses the potential of multiple images depicting the same pedestrian, enabling the re-ID model to excavate comprehensive information from a single image.

\subsection{Multi-view Learning and Knowledge Distillation}
Multi-view learning \cite{Multiview, CGL, EEOS} is a machine learning paradigm that involves using multiple perspectives or views of data to improve the performance of a learning algorithm. It proves particularly valuable in scenarios where a single feature representation may not capture the complete information of the data. For instance, Li et al. \cite{LI} effectively explored the valuable information across various views by combining their respective features into a joint model. Zheng et al. \cite{ZHENG} introduced degeneration mapping models and a low-rank tensor constraint to exploit both complementary and consensus information implied by multi-view data. Tang et al. \cite{MGFWL} developed a multiple graph fusion strategy to learn a unified similarity graph for exploiting sufficient relations of data samples. Considering the original data may not be separable into subspaces, Huang et al. \cite{HUANG} maintained the graph's geometric structure to learn a smooth comprehensive representation. Moreover, He et al. \cite{USSFA} proposed a unified spectral-spatial feature aggregation algorithm to harness the rich spectral and spatial complementary information inherent in the hyperspectral image.

Knowledge distillation \cite{KD} is a machine learning technique that aims to transfer knowledge or expertise from a larger and more complex model, referred to as the teacher model, to a smaller and more compact model, known as the student model. The goal is to enable the student model to achieve comparable performance to the teacher model while being computationally more efficient. Particularly, there has been extensive research on knowledge distillation in the context of person re-ID. Gu et al. \cite{Temporal} addressed the issue of information asymmetry between images and videos by propagating temporal knowledge learned by a video representation network to an image representation network. To tackle the occlusion problem, Kiram et al. \cite{HG} proposed a teacher-student framework that matches the distributions of occluded samples with those of holistic ones. 

Recently, recognizing the rich complementary information present in multi-view pedestrian images, Jin et al. \cite{UMTS} and Angelo et al. \cite{MVKD} combined multi-view learning and knowledge distillation to guide the holistic re-ID model identification. Specifically, they introduced a teacher network that takes multiple images of the same pedestrian as input and utilized the knowledge learned from this network as a supervision signal to train a student network, with a single-view image as input, thereby enhancing the recognition performance of image-to-image and image-to-video person re-ID, respectively. However, these methods mainly focus on holistic person re-ID and encounter challenges in generalizing well to the occluded scenarios. This is attributed to the presence of noisy information and inconsistent discriminative nature across the information from different occluded images. In this paper, our MVI$^{2}$P framework incorporates Localization and Quantification modules, enhancing the accurateness and reliability of the comprehensive information.

\section{The Proposed Method}

\begin{figure}[t!]
  \centering
  \includegraphics[width=4.8in,height=3.4in]{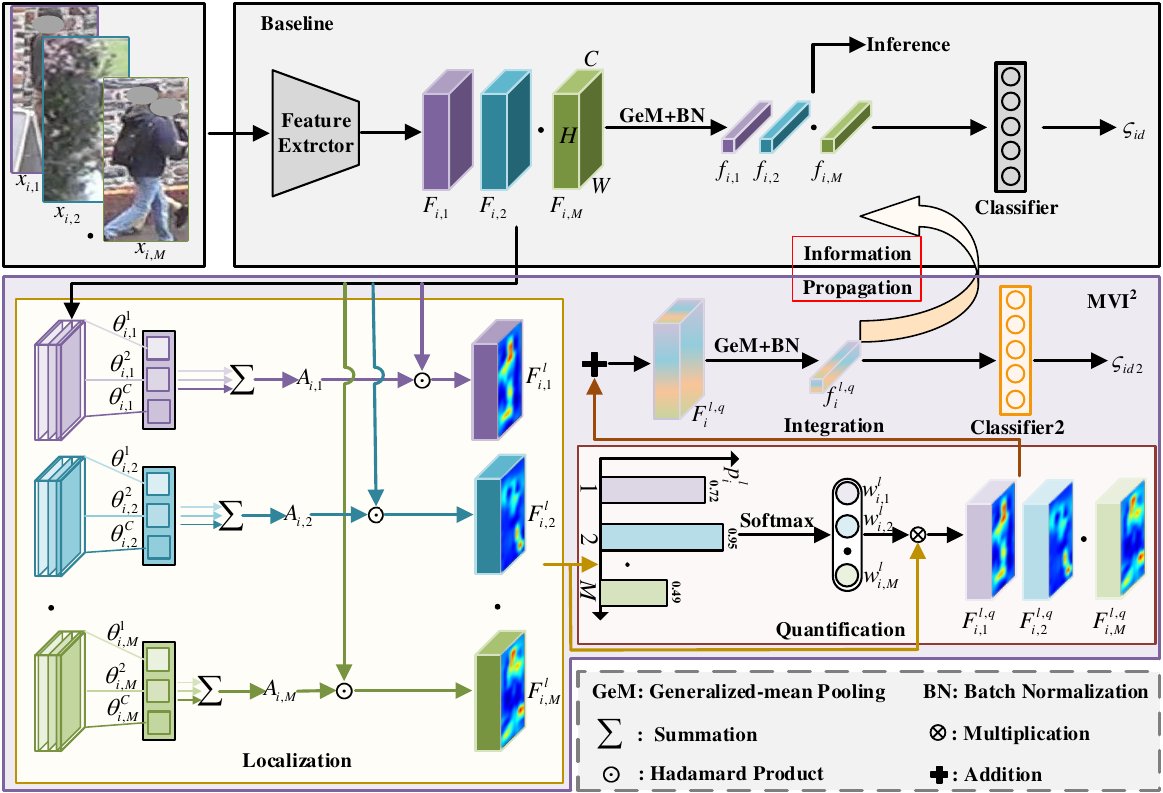}\\
  \caption{Overview of our MVI$^{2}$P framework. Given multiple images with the same identity, MVI$^{2}$P: \textbf{i)Localizes} spatial discriminative regions within feature maps to filter out noisy information. \textbf{ii)Quantifies} the relative importance of saliency information from different images. \textbf{iii) Integrates} multi-view information by performing element-wise addition. \textbf{iv)Propagates} comprehensive information implied by multiple images into a single image through knowledge distillation. During the inference phase, the similarity is measured using the features extracted from the single-view image.}
  \label{Fig:2}
\end{figure}

In this section, we provide a detailed implementation of the developed MVI$^{2}$P, as illustrated in Figure \ref{Fig:2}. Initially, we introduce the Baseline and identify its limitations. Subsequently, we present the Multi-view Information Integration module, encompassing Localization, Quantification, and Integration. Additionally, we elaborate on the Information Propagation mechanism. Finally, we discuss the differences and advantages of our MVI$^{2}$P compared to existing methods.

\subsection{Baseline}

Similar to the majority of mainstream methods, we utilize the standard re-ID model as our baseline branch, comprising a feature extractor and a classifier. In each mini-batch, we randomly sample $P$ identifies and $K$ images per identity. Taking $x_{i,m}$ (the $m$-th image of the $i$-th pedestrian) as an example, the feature extractor is responsible for extracting its feature map $F_{i,m}$. The feature map is then subjected to generalized-mean pooling \cite{AGW} and batch normalization operations. The resulting feature vector $f_{i,m}$ is subsequently passed to the classifier. We employ identity loss to optimize the network:
\begin{equation}
\mathcal{L}_{id}=-log(p(y_{i}|Wf_{i,m})),
\end{equation}
where $y_{i}$ is the ground truth corresponding to the image $x_{i,m}$, and $W$ denotes the parameter of the classifier.

The Baseline branch ensures the discriminative nature of the feature vector utilized for inference. However, it extracts features independently from each image, ignoring the complementary information from others. As a result, it fails to capture the comprehensive characteristics of the target pedestrian. Particularly, in the case of occlusion, the model's attention is restricted to a limited region within the visible areas.

\subsection{Multi-view Information Integration}
During training, multi-view images with the same identity are accessible, denoted as $\{x_{i,1}, x_{i,2}, \cdots, x_{i,M}\}$, and these images can offer rich complementary information. Obviously, integrating their feature maps $\{F_{i,1}, F_{i,2}, \cdots, F_{i,M}\}$ to measure similarity would effectively mitigate the difficulty in identity matching.

However, as previously discussed, not all information contributes to the identification process. In situations where the target pedestrian is occluded, the presence of information from obstacles and non-target pedestrians can impede recognition. Therefore, it is crucial to exercise discretion when integrating information from different images, rather than doing so indiscriminately. This discretion aligns with human cognitive behavior, as humans possess the innate ability to filter out irrelevant information and selectively integrate only what is advantageous for identification. Additionally, different body parts typically exhibit varying levels of discriminative properties. For instance, when distinguishing between pedestrians, facial features often receive more attention than hairstyles. Consequently, it becomes essential to assign distinct weights to different types of information during the integrating process. In this section, we develop the MVI$^{2}$ that incorporates Localization and Quantification modules, and an Integration operation, as illustrated in Figure \ref{Fig:2}.

\subsubsection{Localization}
Inspired by class activation maps (CAMs) \cite{CAM}, which provide an interpretation of which parts of the feature map are most responsible for the model's decision, we present a CAMs-aware Localization module to investigate spatially discriminative regions within the feature map. To be specific, given individual feature map $F_{i,m}\in \mathbb{R}^{H\times W\times C}$ and its corresponding feature vector $f_{i,m}\in \mathbb{R}^{C}$, where $H$, $W$, and $C$ represent height, width, and number of channels, respectively. By acquiring the parameter $\theta_{i,m}\in \mathbb{R}^{C}$ from the classifier, which indicates the weight of categorizing $f_{i,m}$ into the corresponding identity neuron, we can compute the CAMs as follows:
\begin{equation}
A_{i,m}=\sum_{c}\theta_{i,m}^{c}F_{i,m}^{c},
\end{equation}
$A_{i,m}$ enables us to realize which areas in $F_{i,m}$ are contribute to identification. Performing the Hadamard product between $A_{i,m}$ and $F_{i,m}$ yields $F_{i,m}^{l}$ that accentuates significant information:
\begin{equation}
F_{i,m}^{l}=A_{i,m}\odot F_{i,m}.
\end{equation}

\subsubsection{Quantification}
The probability that the feature map is correctly identified by the classifier reveals its discriminative nature and the feature map with a higher discriminative is more important to the identification. Therefore, we develop a probability-aware Quantification module to assess the importance of saliency information present in the feature map. To be specific, upon obtaining the above $F_{i,m}^{l}$, we apply pooling and normalization operations to it. Subsequently, we input the processed feature vector $f_{i,m}^{l}$ into the classifier, resulting in the probability $p_{i,m}^{l}$ of correct classification. This probability serves as an indicator of the discriminative nature exhibited by $F_{i,m}^{l}$. In principle, a higher discriminative feature map should be assigned a greater weight in the process of information integration. We compute the weight as follows:
\begin{equation}
w_{i,m}^{l}=\frac{exp(p_{i,m}^{l})}{\sum_{m=1}^{M}exp(p_{i,m}^{l})},
\end{equation}
$w_{i,m}^{l}$ reflects the relative importance of $F_{i,m}^{l}$. Performing the Multiplication between $w_{i,m}^{l}$ and $F_{i,m}^{l}$ yields $F_{i,m}^{l,q}$ that emphasize the proportional contributions:
\begin{equation}
F_{i,m}^{l,q}=w_{i,m}^{l}\otimes F_{i,m}^{l}.
\end{equation}

\subsubsection{Integration} Accordingly, we integrate information with multiple feature maps by performing element-wise addition:
\begin{equation}
F_{i}^{l,q}=\sum_{m=1}^{M}F_{i,m}^{l,q},
\end{equation}
$F_{i}^{l,q}$ provides a comprehensive representation while filtering noise information and possessing high reliability. Similar to the Baseline, after applying pooling and normalization operations to $F_{i}^{l,q}$,  we enforce an identity loss on the resulting feature vector $f_{i}^{l,q}$ to guarantee its discrimination:

\begin{equation}
\mathcal{L}_{id}=-log(p(y_{i}|Wf_{i}^{l,q})),
\end{equation}
where $W_{2}$ refers to the parameter of the classifier2.

\subsection{Information Propagation}
The MVI$^{2}$ generates a more comprehensive representation of the target pedestrian's characteristics that surpasses the individual feature maps learned from the Baseline branch. This enables more accurate retrieval of pedestrian images of the same identity during inference. However, in the testing phase, only a single image is provided as the query, and the association between identities in different images is unknown. Therefore, it is crucial to develop an effective method to guide the model in reasoning multi-view information from a single-view image.

We present the Information Propagation (IP) mechanism to accomplish the above objective. Specifically, drawing inspiration from knowledge distillation, which facilitates the transfer of knowledge from a larger model to a smaller one, we propagate the comprehensive information encapsulated in the integrated feature vector to the individual feature vector:
\begin{equation}
\mathcal{L}_{kd}=\left\|f_{i}^{l,q}-f_{i,m}\right\|_{2},
\label{eq4}
\end{equation}
where $\left\|\cdot \right\|_{2}$ indicates $l_{2}$-norm.

Trained with the IP mechanism, the feature extractor is capable of learning comprehensive representation from a single-view image, mining more discriminative grounds for identification.

\subsection{Optimization and Inference}

The total training loss can be formulated as:
\begin{equation}
\mathcal{L}_{total}=\mathcal{L}_{id}+\mathcal{L}_{id2}+\lambda\mathcal{L}_{kd},
\end{equation}
where $\lambda$ balances the contribution on $\mathcal{L}_{kd}$.

The entire MVI$^{2}$P framework is trained in an end-to-end manner. During inference, we utilize the Baseline branch to extract the feature vector from each query and gallery image and measure their similarities using cosine distance.

\subsection{Discussion}
In the community of person re-ID, there are two typical works that explore the application of multi-view information in person re ID, UMTS \cite{UMTS} and VKD \cite{MVKD}. However, they primarily target holistic person re-ID task, while the potential of harnessing such information in occluded scenarios has not been thoroughly explored. By comparison, our MVI$^{2}$P framework distinguishes itself from existing approaches by offering the following unique differences and advantages: (Here, we primarily discuss the proposed MVI$^{2}$P and UMTS as VKD is designed for the video task.). 

1). UMTS directly concatenates multi-view images to learn the comprehensive representation, which defaults that all information in each image contributes to the identification. This strategy is feasible for holistic person re-ID but becomes inappropriate in occluded scenarios as the noisy information in occluded images impedes the information integration process. Our Localization efficiently addresses this issue by identifying the spatial discriminative regions of the target pedestrian, allowing us to selectively focus on the relevant information that significantly contributes to accurate recognition. 

2). UMTS concatenates multiple images in an equal manner, defaulting that the information provided by different images contributes equally to the identification. However, it fails to consider a crucial fact: different information may possess varying degrees of reliability. Consequently, it is imperative to devise an effective mechanism that quantifies the proportion of different information during the integration process. Our Quantification successfully fulfills this purpose.

3). UMTS requires a specially designed teacher network to learn a comprehensive representation that incorporates information from multiple images. In contrast, our MVI$^{2}$P eliminates the reliance on a teacher network, resulting in a substantial reduction of half the model parameters.

In summary, the motivation is reasonable and innovative. The challenges addressed are meaningful and significant. Noteworthy, the Localization and Quantification modules do not introduce external networks with parameters. The entire MVI$^{2}$P framework is simple yet effective. We have thoroughly demonstrated these claims through extensive experiments.

\section{Experiments}

\subsection{Datastes}
We conduct experiments on five public person re-ID datasets, including three occluded re-ID datasets (Occluded-DukeMTMC \cite{PGFA}, P-DukeMTMC-reID \cite{OCCLUDEDREID}, and Occluded REID \cite{OCCLUDEDREID}), as well as two widely used holistic re-ID datasets (Market-1501 \cite{Market} and MSMT17 \cite{PTGAN}).

\textbf{Occluded-DukeMTMC} is a challenging dataset specially constructed for occluded person re-ID, where all the samples are derived from DukeMTMC-reID \cite{Duke}. It comprises 15,618 training images of 702 pedestrians, 2,210 query images of 519 pedestrians, and 17,661 gallery images of 1,110 pedestrians.

\textbf{P-DukeMTMC-reID} contains 24,593 images of 1,299 pedestrians, of which 12,927 images of 665 identities are used for training and the rest are used for testing. This dataset is also derived from DukeMTMC-reID, and all query images in the test set are disturbed by different types of occlusions.

\textbf{Occluded-REID} is a small-scale occluded person database that contains 2,000 images of 200 identities. Each pedestrian has five occluded images and five full-body images. According to the protocol in \cite{OCCLUDEDREID, IGOAS}, all images are equally divided into two groups, one for training and the other for testing.

\textbf{Market-1501} is a widely used holistic person dataset captured from 6 non-overlapping camera views. It comprises 12,936 training images featuring 751 distinct identities and 23,100 testing images of 750 pedestrians. All images have been uniformly cropped, with a small portion being subject to occlusion.

\textbf{MSMT17} is currently the largest holistic person database, encompasses 32,621 training images of 1,041 pedestrians, along with 11,659 query images of 3,060 pedestrians, and 82,161 gallery images of 3,060 pedestrians. All images are captured by 15 cameras, including 12 outdoor cameras and 3 indoor cameras.

The above settings of datasets are summarized in Table \ref{Tab:1}.

\begin{table}[!ht]\small
\centering {\caption{Details of person re-ID datasets. 'ID': Number of identities. 'IMGS': Number of images. All datasets are abbreviated.}\label{Tab:1}
\renewcommand\arraystretch{1.0}
\begin{tabular}{c|cc|cc|cc}
\hline
 \hline
  \multirow{2}*{Datasets} & \multicolumn{2}{c|}{Training} & \multicolumn{2}{c|}{Query} & \multicolumn{2}{c}{Gallery}\\
 & ID & Imgs & ID & Imgs & ID & Imgs\\
  \hline
  
  O-Duke & 702 & 15,618 & 519 & 2,210 & 1,110 & 17,661 \\

  P-Duke & 665 & 12,927 & 634 & 2,163 & 634 &9,053 \\

  O-REID  & 100 & 1,000 & 100 & 500 & 100 & 500 \\
  
  Market  & 751 & 12,936 & 750 & 3,368 & 750 & 19,732 \\

  MSMT & 1,041 & 32,621 & 3,060 & 11,659 & 3,060 & 82,161 \\
  
  \hline\hline
\end{tabular}}
\end{table}

\subsection{Settings}
\textbf{Implementation Details}: Our MVI$^{2}$P adopts ResNet-50 as the backbone, which is pre-trained on ImageNet \cite{Imagenet}. All experiments are conducted on a single GTX3090 GPU with the Pytorch platform. Data augmentation techniques, including random flipping, padding, and cropping, are applied, and all images are uniformly resized to 256$\times$128. We apply the Label Smoothing \cite{LS} regularization trick to the identity loss to prevent overfitting. We employ Adam optimizer \cite{Adam} to optimize the model parameters. The batch size is set to 64, with each batch comprising 8 identities, and each identity contains 8 images. The learning rate is initialized to $3\times10^{-4}$ and decreased by a factor of 0.1 at the 40th epoch and 70th epoch, respectively. The model is trained for a total of 120 epochs. The hyper-parameter is set to $\lambda=0.007$. Moreover, $M=4$ indicates integrating information from 4 different images to propagate.

\textbf{Evaluation Protocol}: We evaluate the recognition performance using the standard indicators, namely Cumulative Matching Characteristics (CMC) and mean Average Precision (mAP). All experiments are performed in the single query setting.

\subsection{Comparison with State-of-the-art Methods}
We compare the recognition performance of the developed MVI$^{2}$P with state-of-the-art methods for both occluded and holistic person re-ID tasks.

\begin{table}[!ht]\small
\centering {\caption{Performance comparison with state-of-the-art methods on Occluded-DukeMTMC. 'A' denotes the method based on external model-assisted and 'F' is the method based on external model free. 'Bac' indicates the backbone (Res: ResNet-50. Res-IBN: ResNet-50-IBN. Trans: Transformer). 'God' represents whether the method generates additional occlusion data. '-' denotes that no reported result is available.}\label{Tab:2}
\renewcommand\arraystretch{1.0}
\begin{tabular}{c|c|c|c|cccc}
\hline
 \hline
  & \multirow{2}*{Methods} & \multirow{2}*{Bac} & \multirow{2}*{God} & \multicolumn{4}{c}{Occluded-DukeMTMC} \\
\cline{5-8} & & & & Rank-1 & Rank-5 & Rank-10 & mAP \\
\hline

   \multirow{10}{0.5cm}{\centering A} & PGFA \cite{PGFA} & Res & \XSolidBrush & 51.4 & 68.6 & 74.9 & 37.3 \\

  & HOReID \cite{HOREID} & Res & \XSolidBrush & 55.1 & - & - & 43.8 \\

  & PGMANet \cite{PGMANet} & Res & \XSolidBrush & 51.3 & 66.5 & 73.4 & 40.9 \\
  
  & SORN \cite{SORN} & Res & \XSolidBrush & 57.6 & 73.7 & 79.0 & 46.3 \\

  & Pirt \cite{PIRT} & Res-IBN & \XSolidBrush & 60.0 & - & - & 50.9 \\

  & OAMN \cite{OAMN} & Res & \Checkmark & 62.6 & 77.5 & - & 46.1 \\

   & PGFL-KD \cite{PGFL-KD} & Res & \XSolidBrush & 63.0 & - & - & 54.1 \\

   & RFCnet \cite{RFC} & Res & \XSolidBrush & \underline{63.9} & \underline{77.6} & \underline{82.1} & \underline{54.5} \\

   & SCS \cite{SCS} & Res & \XSolidBrush & 60.5 & 74.6 & 79.3 & 52.2 \\

   & MFEN \cite{MFEN} & Res & \XSolidBrush & 60.8 & - & - & 47.6 \\

  \hline
  
  \multirow{15}{0.5cm}{\centering F} & ISP \cite{ISP} & Res & \XSolidBrush & 62.8 & 78.1 & 82.9 & 52.3\\

  & CBDBNet \cite{CBDBNet} & Res & \XSolidBrush & 50.9 & 66.0 & 74.2 & 38.9 \\

  & ASAN \cite{ASAN} & Res & \Checkmark & 55.4 & 72.4 & 78.9 & 43.8 \\

  & IGOAS \cite{IGOAS} & Res & \Checkmark & 60.1 & - & - & 49.4 \\

  & HG \cite{HG} & Res & \Checkmark & 61.4 & 77.0 & 79.8 & 50.5 \\

  & PAT \cite{PAT} & Trans & \XSolidBrush & 64.5 & - & - & 53.6 \\

  & MHSANet \cite{MHSANET} & Res & \XSolidBrush & 59.7 & 74.3 & 79.5 & 44.8 \\

  & QPM \cite{QPM} & Res & \XSolidBrush & 64.4 & 79.3 & 84.2 & 49.7 \\

  & DRL-Net \cite{DRL-Net} & Trans & \Checkmark & 65.8 & \underline{80.4} & \underline{85.2} & 53.9 \\

  & OPR-DAAO \cite{OPRDAAO} & Res & \Checkmark & 66.2 & 78.4 & 83.9 & 55.4 \\

  & FED \cite{FED} & Trans & \Checkmark & \underline{68.1} & 79.3 & - & \underline{56.4} \\

  & RTGAT \cite{RTGAT} & Res & \XSolidBrush & 61.0 & 69.7 & 73.6 & 50.1 \\
  
  \cline{2-8} & Baseline & Res & \XSolidBrush & 51.3 & 69.0 & 74.2 & 43.9\\

  & \textbf{MVI$^{2}$P} & Res & \XSolidBrush & \bf{68.2} & \bf{82.1} & \bf{86.2} & \bf{55.6}\\

  & \textbf{MVI$^{2}$P*} & Res-IBN & \XSolidBrush & \bf{68.6} & \bf{81.2} & \bf{86.0} & \bf{57.3}\\
  \hline\hline
\end{tabular}}
\end{table}

\textbf{Results on Occluded-DukeMTMC}: We first conduct experiments on Occluded-DukeMTMC and present the results in Table \ref{Tab:2}. From these results, the following conclusions can be drawn: 

1). With the guidance of visual cues, methods based on external model-assisted have achieved relatively satisfactory results. For instance, RFCNet \cite{RFC} gains recognition rates of 63.9\% Rank-1, 77.6\% Rank-5, 82.1\% Rank-10, and 54.5\% mAP. However, these methods have limited performance improvement due to the difficulty of accurately detecting visible areas in complex scenes, particularly where occlusion by other individuals occurs. In contrast, the developed MVI$^{2}$P improves Rank-1 accuracy by approximately 17\% compared to the Baseline, demonstrating its superiority in overcoming the occlusion problem. 

2). Recently, there has been significant attention given to external model-free algorithms. Within this category, the utilization of high-quality local features to assist the learning of global representation has gained popularity and proven effectiveness. For instance, QPM \cite{QPM} achieves Rank-1 accuracy of 64.4\% and mAP of 49.7\%. Our MVI$^{2}$P demonstrates an mAP that is 5.9\% higher than that of QPM, which suggests that it is more advantageous to represent the occluded pedestrian with multi-view information compared to fine-grained information. 

3). Generating additional occluded data to promote the mining of pedestrian visible information is an effective approach. OPR-DAAO \cite{OPRDAAO} provides compelling evidence of this, achieving 66.2\% Rank-1 accuracy and 55.4\% mAP. Nevertheless, the issue of ensuring the authenticity and validity of the generated data is a significant concern that warrants attention. Furthermore, employing additional data to jointly train the model will inevitably result in increased consumption of computing resources. In contrast, our MVI$^{2}$P circumvents this issue and achieves higher recognition performance. 

4). Larger models possess an inherent advantage in effectively characterizing occluded target pedestrians. PAT \cite{PAT}, employing Transformer \cite{Transformers} as the backbone, achieves 64.5\% Rank-1 accuracy and 53.6\% mAP. Nevertheless, it exhibits higher model complexity in contrast to our MVI$^{2}$P, which utilizes ResNet-50 as the backbone. Furthermore, it attains a lower recognition rate compared to ours. This further indicates the substantial advantage of MVI$^{2}$P in learning comprehensive representations. 

5). The state-of-the-art method, FED \cite{FED}, employs Transformer as its backbone and trains with a generated occluded database, achieving a Rank-1 accuracy of 68.1\%, a Rank-5 accuracy of 79.3\%, and a mAP of 56.4\%. Our developed MVI$^{2}$P utilizes ResNet-50 as the backbone and trains with the original dataset, achieving a Rank-1 accuracy comparable to FED. Moreover, our MVI$^{2}$P improves the Rank-5 accuracy from 79.3\% to 82.1\%. Particularly, when equipping ResNet-50-IBN \cite{Res50ibn} into our framework, MVI$^{2}$P* exhibits further improvement in recognition rates, surpassing FED by 0.9\% in mAP accuracy.

In summary, our developed MVI$^{2}$P framework demonstrates substantial advantages in addressing the occlusion problem. This can be attributed to its ability to drive the model to learn a comprehensive representation of occluded target pedestrians. In particular, the developed Localization and Quantization modules further contribute to the discriminativeness and reliability of the comprehensive representation.

\begin{table}[!ht]\small
\centering {\caption{Performance comparison with the state-of-the-art methods on P-DukeMTMC-reID.}\label{Tab:3}
\renewcommand\arraystretch{1.0}
\begin{tabular}{c|cccc}
\hline
 \hline
  \multirow{2}*{Methods} & \multicolumn{4}{c}{P-DukeMTMC-reID} \\
\cline{2-5} & Rank-1 & Rank-5 & Rank-10 & mAP \\
  \hline

  PCB \cite{PCB} & 79.4 & 87.1 & 90.0 & 63.9 \\

  IDE \cite{IDE} & 82.9 & 89.4 & 91.5 & 65.9 \\

  PVPM \cite{PVPM} & 85.1 & 91.3 & 93.3 & 69.9 \\

  RTGAT \cite{RTGAT} & 85.6 & 91.5 & 93.4 & 74.3 \\

  PGFA \cite{PGFA} & 85.7 & 92.0 & 94.2 & 72.4 \\

   ISP \cite{ISP} & \underline{89.0} & \underline{94.1}& \underline{95.3} & \underline{74.7} \\
  
  \hline
  \textbf{MVI$^{2}$P} & \bf{91.5} & \bf{95.1} & \bf{96.3} & \bf{79.0} \\

  \textbf{MVI$^{2}$P*} & \bf{91.9} & \bf{94.4} & \bf{95.3} & \bf{80.9} \\
  \hline\hline
\end{tabular}}
\end{table}

\textbf{Results on P-DukeMTMC-reID}: To further evaluate the superiority of the developed MVI$^{2}$P, we perform experiments on P-DukeMTMC-reID and present the results in Table \ref{Tab:3}. It can be seen that our method surpasses state-of-the-art methods. For example, compared to the optimal ISP \cite{ISP}, which generates pseudo-labels of pose keypoints through clustering to facilitate extraction of pedestrian visible area information. Despite not necessitating the use of external models, ISP still incurs substantial computational overhead. Nonetheless, our MVI$^{2}$P exhibits a significant performance advantage, particularly in the mAP metric, where it outperforms ISP by 4.3\%. It is clear that our method is notably superior to the existing approaches. 

\begin{table}[!ht]\small
\centering {\caption{Performance comparison with the state-of-the-art methods on Occluded-REID. }\label{Tab:4}
\renewcommand\arraystretch{1.0}
\begin{tabular}{c|cccc}
\hline
 \hline
  \multirow{2}*{Methods} & \multicolumn{4}{c}{Occluded-REID} \\
\cline{2-5} & Rank-1 & Rank-5 & Rank-10 & mAP \\
  \hline

  SVDNet \cite{SVDNet} & 63.1 & 85.1 & - & - \\

  Random Erasing \cite{RE} & 65.8 & 87.9 & - & - \\

   PCB \cite{PCB} & 66.6 & 89.2 & - & - \\

  AFPB \cite{OCCLUDEDREID} & 68.1 & 88.3 & \underline{93.7} & - \\

   Teacher-S \cite{Teacher-S} & 73.7 & \underline{92.9} & - & - \\

  IGOAS \cite{IGOAS} & \underline{81.1} & 91.6 & - & - \\
  
  \hline
  \textbf{MVI$^{2}$P} & \bf{85.5} & \bf{93.8} & \bf{96.6} & \bf{77.4} \\

  \textbf{MVI$^{2}$P*} & \bf{88.0} & \bf{95.3} & \bf{97.7} & \bf{81.7} \\
  
  \hline\hline
\end{tabular}}
\end{table}

\textbf{Results on Occluded-REID}: To provide a comprehensive assessment of the proposed method's advantages, we conduct comparative experiments on the Occluded-REID dataset. In accordance with IGOAS \cite{IGOAS},  due to some studies not explicitly specifying the training scheme (whether it is based on Occluded-REID or Market-1501 dataset), we primarily compare our method with state-of-the-art approaches trained on Occluded-REID. Furthermore, to ensure fairness, we compute the average of 10 experimental results. As illustrated in Table \ref{Tab:4}, Teacher-S \cite{Teacher-S} and IGOAS have demonstrated superior recognition performance by training the model jointly with original inputs and simulated occlusion data. In contrast, our MVI$^{2}$P outperforms them in Rank-1 accuracy by 9.5\% and 2.1\%, respectively, without requiring additional samples. This once again reinforces the effectiveness of our method in improving the applicability of person re-ID in occlusion scenarios.

\begin{table}[!ht]\small
\centering {\caption{Performance comparison with state-of-the-art methods on Market-1501 and MSMT17. '*' indicates that the code is re-implemented by ourselves.}\label{Tab:5}
\renewcommand\arraystretch{1.0}
\begin{tabular}{c|cc|cc}
\hline
 \hline
  \multirow{2}*{Methods} & \multicolumn{2}{c|}{Market-1501} & \multicolumn{2}{c}{MSMT17} \\
\cline{2-5} & Rank-1 & mAP & Rank-1 & mAP\\
  \hline

  CamStyle \cite{CamStyle} & 88.1 & 68.7 & - & - \\

  FDGAN \cite{FDGAN} & 90.5 & 77.7 & - & - \\

  PTGAN \cite{PTGAN} & - & - & 68.2 & 40.4 \\

   HA-CNN \cite{HACNN} & 91.2 & 75.7 & - & - \\

    PGFA \cite{PGFA} & 91.2 & 76.8 & - & - \\

   PCB \cite{PCB} & 92.3 & 77.4 & - & - \\

    IGOAS \cite{IGOAS} & 93.4 & 84.1 & - & - \\

   AANet \cite{AANet} & 93.9 & 82.5 & - & - \\

    HOReID \cite{HOREID} & 94.2 & 84.9 & - & - \\

   IANet \cite{IANet} & 94.4 & 83.1 & 75.5 & 46.8\\

   BagTricks \cite{BAGTRICKS} & 94.5 & 85.9 & 74.1 & 50.2 \\

  MHSANet \cite{MHSANET} & 94.6 & 84.0 & - & - \\

  UMTS* \cite{UMTS} & 94.3 & 85.8 & 78.1 & 52.6 \\

  DGNet\cite{DGNet} & 94.8 & \underline{86.0} & 77.2 & 52.3\\

   OSNet \cite{OSNet} & \underline{94.8} & 84.9 & \underline{78.7} & \underline{52.9} \\
   
  \hline
  \textbf{MVI$^{2}$P} & \bf{95.2} & \bf{87.0} & \bf{80.4} & \bf{56.4} \\
  \textbf{MVI$^{2}$P*} & \bf{95.3} & \bf{87.9} & \bf{83.9} & \bf{61.4} \\
  \hline\hline
\end{tabular}}
\end{table}

\textbf{Results on Market-1501 and MSMT17}: A superior occluded person re-ID model should not only demonstrate effectiveness in the occluded task but also achieve satisfactory recognition performance in holistic person re-ID. To this end, we evaluate the scalability of the proposed MVI$^{2}$P by deploying it on Market-1501 and MSMT17. The results, presented in Table \ref{Tab:5}, indicate that our MVI$^{2}$P outperforms state-of-the-art methods. Specifically, on Market-1501, the proposed MVI$^{2}$P achieves 95.2\% Rank-1 accuracy and 87.0\% mAP, which are 0.4\% higher than OSNet \cite{OSNet} and 1.0\% higher than DGNet \cite{DGNet}. These results indicate that our method has considerable advantages in holistic person re-ID, primarily due to the rich complementary information contained in multiple images also aiding in identifying the non-occluded target pedestrian. Additionally, our MVI$^{2}$P achieves the best recognition performance on the challenging holistic person dataset, MSMT17. For instance, our Rank-1 accuracy surpasses OSNet by 1.7\%, and our mAP outperforms it by 0.8\%, further demonstrating the strong scalability of the proposed MVI$^{2}$P.

\begin{table}[!ht]\small
 \centering {\caption{Ablation studies of the proposed MVI$^{2}$P. 'Q': Quantification. 'L': Localization. 'IP': Information Propagation.}\label{Tab:6}
\renewcommand\arraystretch{1.0}
\begin{tabular}{c|ccc|cc}
\hline
 \hline
  \multirow{2}*{Methods} & \multicolumn{3}{c|}{Component} & \multicolumn{2}{c}{Occluded-DukeMTMC} \\
\cline{2-6} & Q & L & IP & Rank-1 & mAP \\
  \hline
  
  Baseline &  &  &  & 51.3 & 43.9 \\

  +IP &  &  & $\checkmark$ &  62.0 & 52.8 \\

  +L & & \checkmark & $\checkmark$ & 64.9 & 54.3 \\

  +Q & $\checkmark$ & \checkmark & $\checkmark$ & 68.2 & 55.6 \\

  \hline\hline
\end{tabular}}
\end{table}

\subsection{Model Analysis}

\begin{figure}[th!]
\centering
{\includegraphics[height=3.5in,width=3.5in,angle=0]{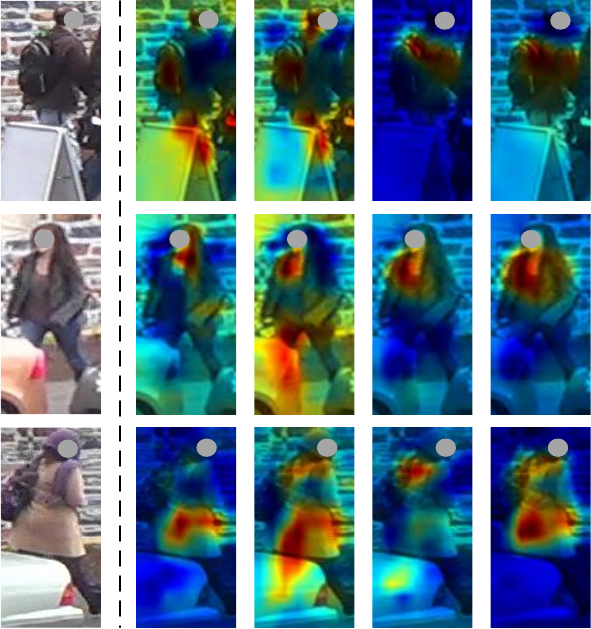}}
\caption{Visualization of the spatial discriminative regions. The images are arranged from left to right in the following order: original image, heatmap obtained by Baseline, IP, IP+L, and IP+L+Q.}
\label{Fig:3}
\end{figure}

\begin{figure}[th!]
\centering
{\includegraphics[height=1.6in,width=4.8in,angle=0]{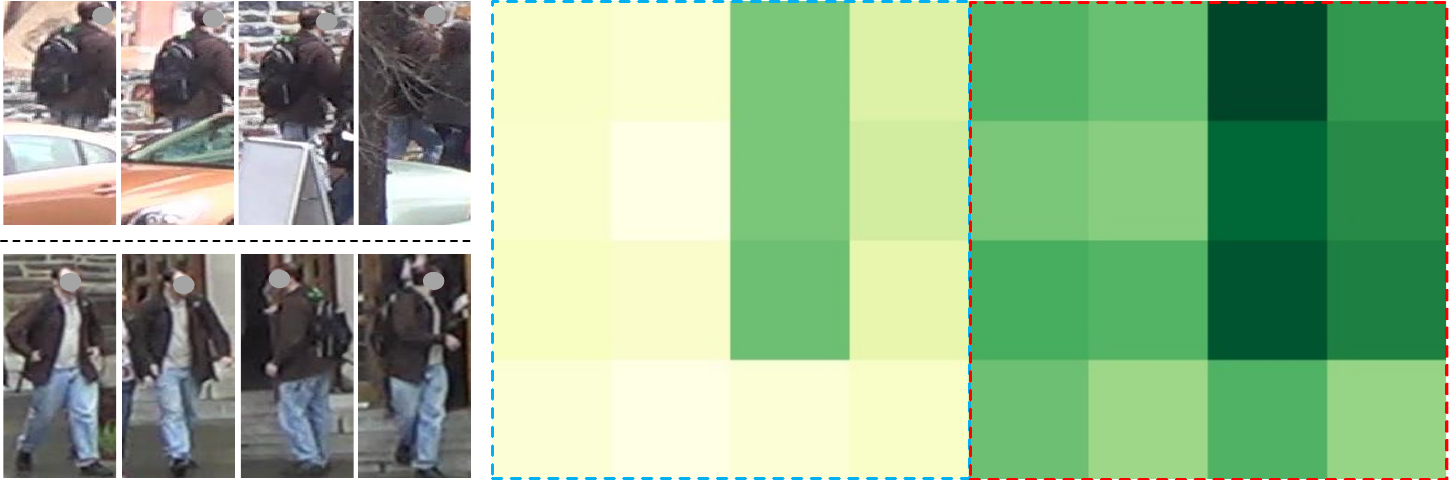}}
\caption{Given the occluded query and holistic gallery images, we extract features from both and calculate their similarities. The results obtained using the Baseline are displayed in the blue box, while the results obtained using our MVI$^{2}$P are shown in the red box. A darker color indicates a higher similarity.}
\label{Fig:4}
\end{figure}

\emph{1) Evaluation on Importance of Multi-view Information}: This paper aims to leverage multi-view information to learn a comprehensive representation of the occluded pedestrian, alleviating the difficulty that the incomplete representation of the single image in matching identity. Table \ref{Tab:6} demonstrates the significant improvements achieved by our IP (integrate information directly and propagate) mechanism, resulting in an 11.7\% increase in Rank-1 accuracy and an 8.9\% increase in mAP compared to the Baseline. Additionally, Figure \ref{Fig:3} reveals that the Baseline model focuses only on a limited subset of visible areas, whereas our IP mechanism encourages the model to consider a broader range. This indicates that more discriminative grounds are explored to facilitate identification. Furthermore, we compute the similarity between the features of the occluded image (query) and the features of the holistic image (gallery). As shown in Figure \ref{Fig:4}, incorporating our IP shortens the distance between the two compared to the Baseline, implying an alleviation of the difficulty in identity matching. These findings provide strong evidence that leveraging multi-view information enables enhancing the model's ability to comprehensively characterize the occluded pedestrian, aligning with our initial motivation.

\emph{2) Contribution of Localization and Quantification}: Our MVI$^{2}$P incorporates two crucial novel ingredients: i) Localization, which allows the model to selectively integrate valuable information for propagation. As illustrated in Table \ref{Tab:6}, the inclusion of Localization results in a 2.9\% improvement in Rank-1 accuracy and a 1.5\% improvement in mAP. Furthermore, Figure \ref{Fig:3} clearly demonstrates that the introduction of this module enables the model to avoid focusing on occlusion noise. ii) Quantification, which is further designed to emphatically integrate high-reliability information. As shown in Table \ref{Tab:7}, when the Quantification is equipped, it leads to a further improvement in Rank-1 accuracy from 64.9\% to 68.2\% and mAP from 54.3\% to 55.6\%. In Particular, Figure \ref{Fig:3} clearly depicts these two modules enabling the model to acquire more comprehensive information. These results provide compelling evidence that the challenges we addressed are reasonable and each component we designed is effective.

\begin{figure}[th!]
\centering
\subfigbottomskip=-1pt
\subfigcapskip=-1pt
\subfigure[$\lambda$] {\includegraphics[height=1.5in,width=2.0in,angle=0]{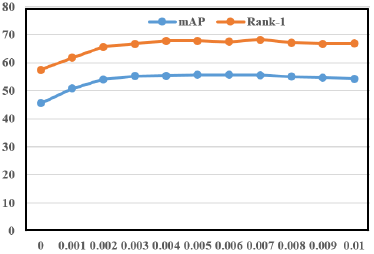}}
\subfigure[$M$] {\includegraphics[height=1.5in,width=2.0in,angle=0]{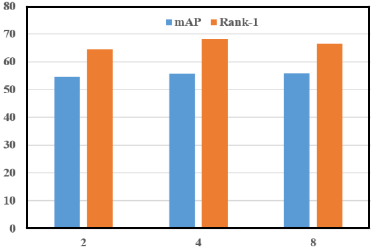}}
\caption{The performance effect analysis on the hyper-parameter $\lambda$ and the number of multi-view images $M$.}
\label{Fig:5}
\end{figure}

\emph{3) Effect of the Hyper-parameter $\lambda$ on Performance}: The MVI$^{2}$P framework involves a hyper-parameter, denoted as $\lambda$, which controls the relative importance of the knowledge distillation loss. The impact of different $\lambda$ values on recognition performance is illustrated in Figure \ref{Fig:5}. Notably, at $\lambda=0.007$, the Rank-1 recognition rate and mAP reach their peak, indicating that 0.007 represents the optimal value for $\lambda$.

\emph{4) Rationality of Integrating Information from 4 Different Images}: We integrate and propagate information from 4 different images to generate a comprehensive representation. To establish its rationality, we conduct experiments using 2, 4, and 8 images, respectively. Figure \ref{Fig:5} demonstrates that integrating information from 4 different images yields the optimal results. Additionally, we observe a slight decrease in performance for $M=2$ and $M=8$. We speculate that the former occurs due to insufficient comprehensive information, while the latter leads to an excessive amount of redundant information.

\subsection{Further Discussion}

\emph{1) Model Complexity}: Our MVI$^{2}$P framework not only demonstrates superior performance but also offers advantages in terms of model complexity. Firstly, unlike methods that rely on external model assistance, our MVI$^{2}$P does not require the introduction of external visual cues, resulting in lower model complexity. Secondly, we observe that current state-of-the-art external model-free methods necessitate high model complexity. For instance, as shown in Table \ref{Tab:7}, IGOAS \cite{IGOAS} employs two feature extraction branches, resulting in 32.6M model parameters. OPR-DAAO \cite{OPRDAAO} utilizes a teacher-student learning approach, requiring 58.2M model parameters. FED \cite{FED} adopts Transformer as the Backbone, demanding 146.2M model parameters. In contrast, our MVI$^{2}$P architecture is relatively simple, employing ResNet-50 as the Backbone and only introducing an additional classifier, requiring only 26.3M model parameters. Furthermore, the aforementioned methods generate additional data for joint training, while ours does not. Additionally, we directly deploy UMTS \cite{UMTS} on the occluded person re-ID task. It can be observed that the number of model parameters is about twice that of ours, but its performance is far behind ours. In summary, besides its performance benefits, our MVI$^{2}$P also possesses a significant advantage in terms of model complexity.

\begin{table}[!t]\small
\centering {\caption{Model complexity analysis. Param: the parameter number of models. ‘God’: generate additional occluded data.}\label{Tab:7}
\renewcommand\arraystretch{1.0}
\begin{tabular}{c|c|c|c|c}
\hline
 \hline
  Methods & Param & God & Rank-1 & mAP \\
  \hline

  IGOAS \cite{IGOAS} & 32.6M & \Checkmark & 60.1 & 49.4 \\

  OPR-DAAO \cite{OPRDAAO} & 58.2M & \Checkmark & 66.2 & 55.4 \\

  FED \cite{FED} & 146.2M & \Checkmark & 68.1 & 56.4 \\

  \hline

  UMTS* \cite{UMTS} & 51.3M & \XSolidBrush & 58.6 & 51.4 \\
  
  \hline
  \textbf{Baseline} & \bf{24.9M} & \XSolidBrush & \bf{51.3} & \bf{43.9} \\
  \textbf{MVI$^{2}$P} & \bf{26.3M} & \XSolidBrush & \bf{68.2} & \bf{55.6} \\
  \textbf{MVI$^{2}$P*} & \bf{26.3M} & \XSolidBrush & \bf{68.6} & \bf{57.3} \\
  \hline\hline
\end{tabular}}
\end{table}

\emph{2) Future Research}: In this paper, we initially explore the potential of leveraging multi-view information to address the occlusion problem. However, certain limitations remain that warrant further attention. Firstly, in our MVI$^{2}$P framework, we integrate multi-view information through element-wise addition. Although reasonable, this approach may not be optimal, as employing an attention mechanism could potentially provide superior information integration capabilities. Secondly, different images may contain partially overlapping visible regions, which could result in an excessive contribution of such shared information during the integration process, potentially limiting performance improvements. These limitations serve as motivation for our future research endeavors.

\section{Conclusion}
In this paper, we propose a Multi-view Information Integration and Propagation (MVI$^{2}$P) framework, driving the model to excavate comprehensive information from a single occluded image. Specifically, MVI$^{2}$P integrates the feature maps of multi-view images to generate a comprehensive representation. To address the challenges posed by occlusion noise and significant differences between different information sources, the CAMs-aware Localization and probability-aware Quantification modules are developed to enhance the discriminative and reliable nature of the representation. Furthermore, since multiple images with the same identity are only available during training, an Information Propagation (IP) mechanism is designed to distill knowledge from the comprehensive representation to that of the single occluded image. Extensive experiments and analyses convincingly demonstrate the effectiveness and superiority of the MVI$^{2}$P framework. Moving forward, we aim to further investigate the potential of multi-view information in object re-identification.




\bibliography{mybibfile}

\begin{thebibliography}{10}
\expandafter\ifx\csname url\endcsname\relax
  \def\url#1{\texttt{#1}}\fi
\expandafter\ifx\csname urlprefix\endcsname\relax\def\urlprefix{URL }\fi
\expandafter\ifx\csname href\endcsname\relax
  \def\href#1#2{#2} \def\path#1{#1}\fi

\bibitem{SHABAN}
M.~Shaban, A.~Mahmood, S.~A. Al-Maadeed, N.~Rajpoot, An information fusion framework for person localization via body pose in spectator crowds, Information Fusion 51 (2019) 178--188.

\bibitem{HIGCN}
R.~Yan, L.~Xie, J.~Tang, X.~Shu, Q.~Tian, Higcin: Hierarchical graph-based cross inference network for group activity recognition, IEEE Transactions on Pattern Analysis and Machine Intelligence 45~(6) (2023) 6955--6968.

\bibitem{CCG-LSTM}
J.~Tang, X.~Shu, R.~Yan, L.~Zhang, Coherence constrained graph lstm for group activity recognition, IEEE Transactions on Pattern Analysis and Machine Intelligence 44~(2) (2022) 636--647.

\bibitem{Yan1}
S.~Yan, H.~Tang, L.~Zhang, J.~Tang, Image-specific information suppression and implicit local alignment for text-based person search, IEEE Transactions on Neural Networks and Learning Systems (2023) 1--14\href {http://dx.doi.org/10.1109/TNNLS.2023.3310118} {\path{doi:10.1109/TNNLS.2023.3310118}}.

\bibitem{Yan2}
S.~Yan, N.~Dong, L.~Zhang, J.~Tang, Clip-driven fine-grained text-image person re-identification, IEEE Transactions on Image Processing (2023) 1--1\href {http://dx.doi.org/10.1109/TIP.2023.3327924} {\path{doi:10.1109/TIP.2023.3327924}}.

\bibitem{AN}
F.-P. An, J.~e~Liu, Pedestrian re-identification algorithm based on visual attention-positive sample generation network deep learning model, Information Fusion 86-87 (2022) 136--145.

\bibitem{SUUTALA}
Methods for person identification on a pressure-sensitive floor: Experiments with multiple classifiers and reject option, Information Fusion 9~(1) (2008) 21--40, special Issue on Applications of Ensemble Methods.

\bibitem{AGW}
M.~Ye, J.~Shen, G.~Lin, T.~Xiang, L.~Shao, S.~C.~H. Hoi, Deep learning for person re-identification: A survey and outlook, IEEE Transactions on Pattern Analysis and Machine Intelligence 44~(6) (2022) 2872--2893.

\bibitem{BAGTRICKS}
H.~Luo, Y.~Gu, X.~Liao, S.~Lai, W.~Jiang, Bag of tricks and a strong baseline for deep person re-identification, in: Proceedings of the IEEE Conference on Computer Vision and Pattern Recognition Workshops (CVPRW), 2019, pp. 1487--1495.

\bibitem{OCCLUDEDREID}
J.~Zhuo, Z.~Chen, J.~Lai, G.~Wang, Occluded person re-identification, in: 2018 IEEE International Conference on Multimedia and Expo (ICME), 2018, pp. 1--6.

\bibitem{PGFA}
J.~Miao, Y.~Wu, P.~Liu, Y.~Ding, Y.~Yang, Pose-guided feature alignment for occluded person re-identification, in: 2019 IEEE/CVF International Conference on Computer Vision (ICCV), 2019, pp. 542--551.

\bibitem{HOREID}
G.~Wang, S.~Yang, H.~Liu, Z.~Wang, Y.~Yang, S.~Wang, G.~Yu, E.~Zhou, J.~Sun, High-order information matters: Learning relation and topology for occluded person re-identification, in: 2020 IEEE/CVF Conference on Computer Vision and Pattern Recognition (CVPR), 2020, pp. 6448--6457.

\bibitem{OPRDAAO}
S.~Wang, R.~Liu, H.~Li, G.~Qi, Z.~Yu, Occluded person re-identification via defending against attacks from obstacles, IEEE Transactions on Information Forensics and Security 18 (2023) 147--161.

\bibitem{ELF}
D.~Gray, H.~Tao, Viewpoint invariant pedestrian recognition with an ensemble of localized features, in: Computer Vision -- ECCV 2008, 2008, pp. 262--275.

\bibitem{SCN}
Y.~Yang, J.~Yang, J.~Yan, S.~Liao, D.~Yi, S.~Z. Li, Salient color names for person re-identification, in: Computer Vision -- ECCV 2014, 2014, pp. 536--551.

\bibitem{GOG}
T.~Matsukawa, T.~Okabe, E.~Suzuki, Y.~Sato, Hierarchical gaussian descriptor for person re-identification, in: 2016 IEEE Conference on Computer Vision and Pattern Recognition (CVPR), 2016, pp. 1363--1372.

\bibitem{AIESS}
H.~Li, S.~Yan, Z.~Yu, D.~Tao, Attribute-identity embedding and self-supervised learning for scalable person re-identification, IEEE Transactions on Circuits and Systems for Video Technology 30~(10) (2020) 3472--3485.

\bibitem{DML}
D.~Yi, Z.~Lei, S.~Liao, S.~Z. Li, Deep metric learning for person re-identification, in: 2014 22nd International Conference on Pattern Recognition, 2014, pp. 34--39.

\bibitem{AANet}
C.-P. Tay, S.~Roy, K.-H. Yap, Aanet: Attribute attention network for person re-identifications, in: 2019 IEEE/CVF Conference on Computer Vision and Pattern Recognition (CVPR), 2019, pp. 7127--7136.

\bibitem{CamStyle}
Z.~Zhong, L.~Zheng, Z.~Zheng, S.~Li, Y.~Yang, Camstyle: A novel data augmentation method for person re-identification, IEEE Transactions on Image Processing 28~(3) (2019) 1176--1190.

\bibitem{TANG1}
H.~Tang, C.~Yuan, Z.~Li, J.~Tang, Learning attention-guided pyramidal features for few-shot fine-grained recognition, Pattern Recognition 130 (2022) 108792.

\bibitem{TANG2}
H.~Tang, Z.~Li, Z.~Peng, J.~Tang, Blockmix: meta regularization and self-calibrated inference for metric-based meta-learning, in: Proceedings of the 28th ACM international conference on multimedia, 2020, pp. 610--618.

\bibitem{Boosting}
Z.~Zha, H.~Tang, Y.~Sun, J.~Tang, Boosting few-shot fine-grained recognition with background suppression and foreground alignment, IEEE Transactions on Circuits and Systems for Video Technology.

\bibitem{KGS}
Z.~Li, H.~Tang, Z.~Peng, G.-J. Qi, J.~Tang, Knowledge-guided semantic transfer network for few-shot image recognition, IEEE Transactions on Neural Networks and Learning Systems.

\bibitem{PLMSA}
Y.~Wang, J.~Peng, H.~Wang, M.~Wang, Progressive learning with multi-scale attention network for cross-domain vehicle re-identification, Science China Information Sciences 65~(6) (2022) 160103.

\bibitem{CycleGAN}
J.-Y. Zhu, T.~Park, P.~Isola, A.~A. Efros, Unpaired image-to-image translation using cycle-consistent adversarial networks, in: 2017 IEEE International Conference on Computer Vision (ICCV), 2017, pp. 2242--2251.

\bibitem{PGFL-KD}
K.~Zheng, C.~Lan, W.~Zeng, J.~Liu, Z.~Zhang, Z.-J. Zha, Pose-guided feature learning with knowledge distillation for occluded person re-identification, in: Proceedings of the 29th ACM International Conference on Multimedia, 2021, p. 4537–4545.

\bibitem{SORN}
X.~Zhang, Y.~Yan, J.-H. Xue, Y.~Hua, H.~Wang, Semantic-aware occlusion-robust network for occluded person re-identification, IEEE Transactions on Circuits and Systems for Video Technology 31~(7) (2021) 2764--2778.

\bibitem{PIRT}
Z.~Ma, Y.~Zhao, J.~Li, Pose-guided inter- and intra-part relational transformer for occluded person re-identification, 2021, p. 1487–1496.

\bibitem{OAMN}
P.~Chen, W.~Liu, P.~Dai, J.~Liu, Q.~Ye, M.~Xu, Q.~Chen, R.~Ji, Occlude them all: Occlusion-aware attention network for occluded person re-id, in: 2021 IEEE/CVF International Conference on Computer Vision (ICCV), 2021, pp. 11813--11822.

\bibitem{ISP}
K.~Zhu, H.~Guo, Z.~Liu, M.~Tang, J.~Wang, Identity-guided human semantic parsing for person re-identification, in: Computer Vision -- ECCV 2020, 2020, pp. 346--363.

\bibitem{ASAN}
H.~Jin, S.~Lai, X.~Qian, Occlusion-sensitive person re-identification via attribute-based shift attention, IEEE Transactions on Circuits and Systems for Video Technology 32~(4) (2022) 2170--2185.

\bibitem{MHSANET}
H.~Tan, X.~Liu, B.~Yin, X.~Li, Mhsa-net: Multihead self-attention network for occluded person re-identification, IEEE Transactions on Neural Networks and Learning Systems (2022) 1--15.

\bibitem{QPM}
P.~Wang, C.~Ding, Z.~Shao, Z.~Hong, S.~Zhang, D.~Tao, Quality-aware part models for occluded person re-identification, IEEE Transactions on Multimedia (2022) 1--1.

\bibitem{IGOAS}
C.~Zhao, X.~Lv, S.~Dou, S.~Zhang, J.~Wu, L.~Wang, Incremental generative occlusion adversarial suppression network for person reid, IEEE Transactions on Image Processing 30 (2021) 4212--4224.

\bibitem{Multiview}
C.~Xu, D.~Tao, C.~Xu, A survey on multi-view learning, arXiv preprint arXiv:1304.5634.

\bibitem{CGL}
Z.~Li, C.~Tang, X.~Liu, X.~Zheng, W.~Zhang, E.~Zhu, Consensus graph learning for multi-view clustering, IEEE Transactions on Multimedia 24 (2022) 2461--2472.
\newblock \href {http://dx.doi.org/10.1109/TMM.2021.3081930} {\path{doi:10.1109/TMM.2021.3081930}}.

\bibitem{EEOS}
J.~Wang, C.~Tang, Z.~Wan, W.~Zhang, K.~Sun, A.~Y. Zomaya, Efficient and effective one-step multiview clustering, IEEE Transactions on Neural Networks and Learning Systems (2023) 1--12\href {http://dx.doi.org/10.1109/TNNLS.2023.3253246} {\path{doi:10.1109/TNNLS.2023.3253246}}.

\bibitem{LI}
J.~Li, B.~Zhang, G.~Lu, D.~Zhang, Generative multi-view and multi-feature learning for classification, Information Fusion 45 (2019) 215--226.

\bibitem{ZHENG}
Q.~Zheng, J.~Zhu, Z.~Li, Z.~Tian, C.~Li, Comprehensive multi-view representation learning, Information Fusion 89 (2023) 198--209.

\bibitem{MGFWL}
C.~Tang, X.~Zheng, W.~Zhang, X.~Liu, X.~Zhu, E.~Zhu, Unsupervised feature selection via multiple graph fusion and feature weight learning, Science China Information Sciences 66~(5) (2023) 1--17.

\bibitem{HUANG}
S.~Huang, Y.~Liu, H.~Cai, Y.~Tan, C.~Tang, J.~Lv, Smooth representation learning from multi-view data, Information Fusion (2023) 101916.

\bibitem{USSFA}
X.~He, C.~Tang, X.~Liu, W.~Zhang, K.~Sun, J.~Xu, Object detection in hyperspectral image via unified spectral–spatial feature aggregation, IEEE Transactions on Geoscience and Remote Sensing 61 (2023) 1--13.
\newblock \href {http://dx.doi.org/10.1109/TGRS.2023.3307288} {\path{doi:10.1109/TGRS.2023.3307288}}.

\bibitem{KD}
G.~Hinton, O.~Vinyals, J.~Dean, Distilling the knowledge in a neural network, arXiv preprint arXiv:1503.02531.

\bibitem{Temporal}
X.~Gu, B.~Ma, H.~Chang, S.~Shan, X.~Chen, Temporal knowledge propagation for image-to-video person re-identification, in: Proceedings of the IEEE International Conference on Computer Vision (ICCV), 2019, pp. 9647--9656.

\bibitem{HG}
M.~Kiran, R.~G. Praveen, L.~T. Nguyen-Meidine, S.~Belharbi, L.-A. Blais-Morin, E.~Granger, Holistic guidance for occluded person re-identification, arXiv preprint arXiv:2104.06524.

\bibitem{UMTS}
X.~Jin, C.~Lan, W.~Zeng, Z.~Chen, Uncertainty-aware multi-shot knowledge distillation for image-based object re-identification, in: Proceedings of the AAAI Conference on Artificial Intelligence, Vol.~34, 2020, pp. 11165--11172.

\bibitem{MVKD}
A.~Porrello, L.~Bergamini, S.~Calderara, Robust re-identification by multiple views knowledge distillation, in: Proceedings of the European Conference on Computer Vision (ECCV), Springer, 2020, pp. 93--110.

\bibitem{CAM}
B.~Zhou, A.~Khosla, A.~Lapedriza, A.~Oliva, A.~Torralba, Learning deep features for discriminative localization, in: Proceedings of the IEEE Conference on Computer Vision and Pattern Recognition (CVPR), 2016, pp. 2921--2929.

\bibitem{Market}
L.~Zheng, L.~Shen, L.~Tian, S.~Wang, J.~Wang, Q.~Tian, Scalable person re-identification: A benchmark, in: Proceedings of the IEEE International Conference on Computer Vision (ICCV), 2015, pp. 1116--1124.

\bibitem{PTGAN}
L.~Wei, S.~Zhang, W.~Gao, Q.~Tian, Person transfer gan to bridge domain gap for person re-identification, in: Proceedings of the IEEE Conference on Computer Vision and Pattern Recognition (CVPR), 2018, pp. 79--88.

\bibitem{Duke}
E.~Ristani, F.~Solera, R.~Zou, R.~Cucchiara, C.~Tomasi, Performance measures and a data set for multi-target, multi-camera tracking, in: Proceedings of the European Conference on Computer Vision (ECCV), 2016, pp. 17--35.

\bibitem{Imagenet}
J.~Deng, W.~Dong, R.~Socher, L.-J. Li, K.~Li, L.~Fei-Fei, Imagenet: A large-scale hierarchical image database, in: Proceedings of the IEEE Conference on Computer Vision and Pattern Recognition (CVPR), 2009, pp. 248--255.

\bibitem{LS}
C.~Szegedy, V.~Vanhoucke, S.~Ioffe, J.~Shlens, Z.~Wojna, Rethinking the inception architecture for computer vision, in: Proceedings of the IEEE Conference on Computer Vision and Pattern Recognition (CVPR), 2016, pp. 2818--2826.

\bibitem{Adam}
D.~P. Kingma, J.~Ba, Adam: A method for stochastic optimization, arXiv preprint arXiv:1412.6980.

\bibitem{PGMANet}
Y.~Zhai, X.~Han, W.~Ma, X.~Gou, G.~Xiao, Pgmanet: Pose-guided mixed attention network for occluded person re-identification, in: 2021 International Joint Conference on Neural Networks (IJCNN), IEEE, 2021, pp. 1--8.

\bibitem{RFC}
R.~Hou, B.~Ma, H.~Chang, X.~Gu, S.~Shan, X.~Chen, Feature completion for occluded person re-identification, IEEE Transactions on Pattern Analysis and Machine Intelligence 44~(9) (2022) 4894--4912.

\bibitem{SCS}
K.~Zhu, H.~Guo, S.~Liu, J.~Wang, M.~Tang, Learning semantics-consistent stripes with self-refinement for person re-identification, IEEE Transactions on Neural Networks and Learning Systems (2022) 1--12.

\bibitem{MFEN}
Z.~Liu, Q.~Wang, M.~Wang, Y.~Zhao, Occluded person re-identification with pose estimation correction and feature reconstruction, IEEE Access 11 (2023) 14906--14914.

\bibitem{CBDBNet}
H.~Tan, X.~Liu, Y.~Bian, H.~Wang, B.~Yin, Incomplete descriptor mining with elastic loss for person re-identification, IEEE Transactions on Circuits and Systems for Video Technology 32~(1) (2021) 160--171.

\bibitem{PAT}
Y.~Li, J.~He, T.~Zhang, X.~Liu, Y.~Zhang, F.~Wu, Diverse part discovery: Occluded person re-identification with part-aware transformer, in: 2021 IEEE/CVF Conference on Computer Vision and Pattern Recognition (CVPR), 2021, pp. 2897--2906.

\bibitem{DRL-Net}
M.~Jia, X.~Cheng, S.~Lu, J.~Zhang, Learning disentangled representation implicitly via transformer for occluded person re-identification, IEEE Transactions on Multimedia 25 (2023) 1294--1305.

\bibitem{FED}
Z.~Wang, F.~Zhu, S.~Tang, R.~Zhao, L.~He, J.~Song, Feature erasing and diffusion network for occluded person re-identification, in: Proceedings of the IEEE/CVF Conference on Computer Vision and Pattern Recognition, 2022, pp. 4754--4763.

\bibitem{RTGAT}
M.~Huang, C.~Hou, Q.~Yang, Z.~Wang, Reasoning and tuning: Graph attention network for occluded person re-identification, IEEE Transactions on Image Processing 32 (2023) 1568--1582.

\bibitem{Transformers}
A.~Dosovitskiy, L.~Beyer, A.~Kolesnikov, D.~Weissenborn, X.~Zhai, T.~Unterthiner, Transformers for image recognition at scale, arXiv preprint arXiv:2010.11929.

\bibitem{Res50ibn}
X.~Pan, P.~Luo, J.~Shi, X.~Tang, Two at once: Enhancing learning and generalization capacities via ibn-net, in: Proceedings of the European Conference on Computer Vision (ECCV), 2018, pp. 464--479.

\bibitem{PCB}
Y.~Sun, L.~Zheng, Y.~Yang, Q.~Tian, S.~Wang, Beyond part models: Person retrieval with refined part pooling (and a strong convolutional baseline), in: Proceedings of the European Conference on Computer Vision (ECCV), 2018, pp. 480--496.

\bibitem{IDE}
L.~Zheng, H.~Zhang, S.~Sun, M.~Chandraker, Y.~Yang, Q.~Tian, Person re-identification in the wild, in: Proceedings of the IEEE Conference on Computer Vision and Pattern Recognition (CVPR), 2017, pp. 1367--1376.

\bibitem{PVPM}
S.~Gao, J.~Wang, H.~Lu, Z.~Liu, Pose-guided visible part matching for occluded person reid, in: Proceedings of the IEEE Conference on Computer Vision and Pattern Recognition (CVPR), 2020, pp. 11744--11752.

\bibitem{SVDNet}
H.~Huang, X.~Chen, K.~Huang, Human parsing based alignment with multi-task learning for occluded person re-identification, in: 2020 IEEE International Conference on Multimedia and Expo (ICME), IEEE, 2020, pp. 1--6.

\bibitem{RE}
Z.~Zhong, L.~Zheng, G.~Kang, S.~Li, Y.~Yang, Random erasing data augmentation, in: Proceedings of the AAAI Conference on Artificial Intelligence, Vol.~34, 2020, pp. 13001--13008.

\bibitem{Teacher-S}
J.~Zhuo, J.~Lai, C.~Peijia, A novel teacher-student learning framework for occluded person re-identification, arXiv preprint arXiv:1907.03253.

\bibitem{FDGAN}
Y.~Ge, Z.~Li, H.~Zhao, G.~Yin, S.~Yi, X.~Wang, et~al., Fd-gan: Pose-guided feature distilling gan for robust person re-identification, Advances in Neural Information Processing Systems 31.

\bibitem{HACNN}
W.~Li, X.~Zhu, S.~Gong, Harmonious attention network for person re-identification, in: Proceedings of the IEEE International Conference on Computer Vision (ICCV), 2018, pp. 2285--2294.

\bibitem{IANet}
R.~Hou, B.~Ma, H.~Chang, X.~Gu, S.~Shan, X.~Chen, Interaction-and-aggregation network for person re-identification, in: Proceedings of the IEEE Conference on Computer Vision and Pattern Recognition (CVPR), 2019, pp. 9317--9326.

\bibitem{DGNet}
Z.~Zheng, X.~Yang, Z.~Yu, L.~Zheng, Y.~Yang, J.~Kautz, Joint discriminative and generative learning for person re-identification, in: Proceedings of the IEEE Conference on Computer Vision and Pattern Recognition (CVPR), 2019, pp. 2138--2147.

\bibitem{OSNet}
K.~Zhou, Y.~Yang, A.~Cavallaro, T.~Xiang, Omni-scale feature learning for person re-identification, in: Proceedings of the IEEE Conference on Computer Vision and Pattern Recognition (CVPR), 2019, pp. 3702--3712.

\end{thebibliography}

\end{document}